\documentclass{article}

\pdfoutput=1

\DeclareUnicodeCharacter{00A0}{ }  
\DeclareUnicodeCharacter{202F}{ }  
\DeclareUnicodeCharacter{200B}{}   
\DeclareUnicodeCharacter{2212}{-}  
\DeclareUnicodeCharacter{2010}{-}  
\DeclareUnicodeCharacter{2011}{-}  
\DeclareUnicodeCharacter{2013}{--} 
\DeclareUnicodeCharacter{2014}{---}
\DeclareUnicodeCharacter{2018}{'}  
\DeclareUnicodeCharacter{2019}{'}  
\DeclareUnicodeCharacter{201C}{``} 
\DeclareUnicodeCharacter{201D}{''} 
\DeclareUnicodeCharacter{2009}{ }
\DeclareUnicodeCharacter{200A}{ }

\usepackage{arxiv}
\makeatletter
\providecommand{\headeright}{}
\makeatother

\usepackage[utf8]{inputenc} 
\usepackage[T1]{fontenc}    
\usepackage{hyperref}       
\usepackage{url}            
\usepackage{booktabs}       
\usepackage{amsfonts}       
\usepackage{nicefrac}       
\usepackage{amsmath}        
\usepackage{microtype}      
\usepackage{textgreek}      
\usepackage{graphicx}
\usepackage{doi}
\usepackage{apacite}

\providecommand{\keywords}[1]{\par\noindent\textbf{Keywords: }#1}

\usepackage{titlesec}
\titlespacing{\section}{0pt}{\parskip}{-\parskip}
\titlespacing{\subsection}{0pt}{\parskip}{-\parskip}
\titlespacing{\subsubsection}{0pt}{\parskip}{-\parskip}

\title{Revisiting Training Scale: An Empirical Study of Token Count, Power Consumption, and Parameter Efficiency}

\author{ \href{https://orcid.org/0009-0005-5262-1583}{\includegraphics[scale=0.06]{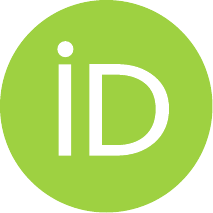}\hspace{1mm}Joe Dwyer}\thanks{I would like to acknowledge the contributions of Jason M. Pittman. His advice and knowledge throughout our time working together has been instrumental to my success.} \\
    Department of Computer Information Science\\
    ECPI University\\
    Virginia Beach, VA 23462 \\
    \texttt{jdwyer@ecpi.edu} \\
}

\hypersetup{
pdftitle={Follow Up Study: Power Consumption},
pdfsubject={q-bio.NC, q-bio.QM},
pdfauthor={Joe Dwyer},
pdfkeywords={LLM Training, Parameter Efficiency, Compute Overhead},
}

\begin{document}
\maketitle

\begin{abstract}
Research in machine learning has questioned whether increases in training token counts reliably produce proportional performance gains in large language models. Building on prior work introducing an energy-aware parameter efficiency metric, this study empirically examines the effects of increasing training token counts under fixed hardware and training conditions. The significance of this work lies in the explicit integration of power consumption and execution duration, as reflected by the power sampling frequency, into token-scale analysis. This addresses a gap in prior studies emphasizing performance outcomes while under-representing computational and energy costs. Using a repeated-measures experimental design on a constant GPU instance with an identical model architecture, optimizer settings, and epoch counts, a 1.1-billion-parameter TinyLlama model was trained at three token counts (500K, 1M, and 2M). While conventional performance metrics exhibited inconsistent or diminishing returns across token scales, the inclusion of power consumption and execution duration revealed a strictly monotonic decline in training efficiency as token count increased. Repeated-measures ANOVA demonstrated a strong effect of token count on parameter efficiency (F(2, 98) = 24,268.23, $p < .001$, $\eta_g^2 = .997$), with all pairwise comparisons remaining significant following the Bonferroni correction. These findings indicate that increases in training token counts may be energetically inefficient even when marginal performance improvements are observed, underscoring the importance of efficiency-aware evaluation in large language model training.
\end{abstract}


\section{Introduction}
This study builds upon prior research examining parameter efficiency in TinyLlama training under varying training token counts (Dwyer, 2025). The underlying parameter efficiency formulation remains consistent with the original study, with the present work extending it to incorporate direct energy measurement. While the original study demonstrated that increases in training token counts did not consistently yield proportional gains in parameter efficiency, it did not explicitly account for power consumption as a contributing factor to compute cost. This omission represents a critical gap, given the increasing importance of energy consumption in evaluating token-based scaling strategies for large language models.

The present study directly extends the dissertation study by operationalizing one of its recommended future research directions: incorporating energy efficiency into the parameter-efficiency framework. The dissertation established a repeated-measures token-scaling design using TinyLlama 1.1B across 500K, 1M, and 2M token conditions, demonstrating that token count significantly affected parameter efficiency under constrained training conditions (Dwyer, 2025). Chapter 5 specifically recommended refactoring the parameter-efficiency formula to incorporate energy efficiency, including performance per TFLOPS per parameter per watt. The present follow-up study implements that recommendation by integrating direct root mean square (RMS) GPU power-consumption measurements into the efficiency calculation, thereby shifting the analysis from compute-aware evaluation to energy-aware evaluation.

\begin{equation}
\mathrm{PE}_{\text{dissertation}}
=
\frac{\mathrm{invPPL}}
{\mathrm{TFLOPS}_{\text{measured}} \cdot \mathrm{MS}_{\text{params}}}
\end{equation}

\begin{equation}
\mathrm{invPPL}
=
\frac{1}{\exp(\mathcal{L}_{\text{eval}})}
\end{equation}

The problem addressed in this research remains the increasing compute overhead associated with linearly scaled training token counts during large language model training, which limits scalability and sustainability (Chennareddy et al., 2024). Accordingly, the purpose of this quantitative experimental study was to explore the effects of varying training token counts on parameter efficiency while explicitly monitoring and incorporating power consumption under fixed model size and tightly controlled training conditions. Statistical significance evaluations remained consistent with the use of repeated-measures ANOVA and post hoc testing, when appropriate.

The present follow-up study extends the original parameter efficiency framework by incorporating direct power consumption measurements collected during training. Power usage, recorded in watts at regular intervals, was integrated into the efficiency calculation using a root mean square formulation, enabling a more comprehensive representation of training cost that reflects both computational and energetic constraints. By explicitly modeling power consumption, this study advances token-scale efficiency analysis beyond performance-centric evaluation. As in previous research, inverse perplexity is the sole performance signal, while parameter count, total token exposure, and measured RMS power consumption define training cost; the TFLOPS term serves as a normalization constant that converts power draw into effective compute throughput.

\begin{equation}
\mathrm{PE}
=
K \cdot
\frac{
\mathrm{invPPL} \cdot \mathrm{CS}_{\mathrm{TFLOPS}} \cdot \mathrm{TT}_{\mathrm{scale}}
}{
\mathrm{MS}_{\mathrm{params}} \cdot \mathrm{TT}_{\mathrm{tokens}} \cdot \mathrm{RMS}(W)
}
\end{equation}

\begin{equation}
\mathrm{RMS}(W)
=
\sqrt{\frac{1}{n}\sum_{i=1}^{n} W_i^2}
\end{equation}

\section{Related Works}
\subsection{Neural Scaling Laws and Compute-Optimal Token Scaling}
The relationship between training scale and model performance has been a central topic in contemporary machine learning research. Early and influential work on neural scaling laws demonstrated predictable improvements in model loss as a function of increased parameters, dataset size, and compute (Kaplan et al., 2020; Brown et al., 2020). Subsequent refinements by Hoffmann et al. (2022) demonstrated that optimal performance depends on balanced scaling between model size and training tokens rather than indiscriminate increases. Complementary theoretical work has explored why scaling laws emerge and how they generalize across regimes (Bahri et al., 2021; Maloney et al., 2022; Boopathy \& Fiete, 2024).

\subsection{Performance-Centric Evaluation and Inverse Perplexity}
Scaling analyses have often emphasized loss-based or perplexity-based metrics, which provide useful information about model behavior but do not fully represent the computational cost of achieving that behavior. In the present study, inverse perplexity is used as the sole performance signal because it converts post-training evaluation loss into a positive efficiency-oriented quantity, where larger values indicate better language-modeling performance. This formulation allows model quality to be placed in the numerator of the efficiency metric while keeping compute, parameter count, token exposure, and power consumption in the denominator. The intent is not to replace task-specific evaluation, but to provide a reproducible and lightweight performance signal suitable for controlled training trials.

\subsection{Compute Cost, Systems Overhead, and Energy Measurement}
Despite advances in scaling law research, scaling efficiency is often evaluated solely in terms of performance outcomes, without explicit consideration of execution duration, power consumption, or system-level overhead. At the systems level, training frameworks have shown that throughput and scalability are shaped not only by algorithms but also by communication and orchestration overhead (Narayanan et al., 2021; Smith et al., 2022; Leung, 2022). However, these analyses typically rely on indirect proxies for cost rather than direct measurement of energy. This limitation motivates the present study's use of direct GPU power readings as a first-class component of the parameter-efficiency framework.

\subsection{Data-Centric and Diminishing-Return Perspectives}
More recent work has questioned whether continued increases in training scale reliably produce proportional gains. Studies examining data limits suggest that scaling may encounter constraints independent of compute availability (Villalobos et al., 2022). Data-centric approaches, such as pruning, have demonstrated that improved outcomes can sometimes be achieved without increasing training volume (Sorscher et al., 2022). Public reporting reflects industry concern that the prevailing scaling strategies may be approaching practical limits (Hu \& Tong, 2024).

\subsection{Positioning of the Present Study}
The study situates itself at the intersection of these lines of work. Rather than proposing new scaling laws, it empirically examines how training token count affects efficiency outcomes under fixed hardware and training conditions. By holding model architecture, optimizer configuration, epoch count, and hardware resources constant, this study isolates the effects of token scaling in a controlled setting. In doing so, it complements prior scaling law research by demonstrating how efficiency-aware evaluation alters the interpretation of training outcomes when increased scale does not yield proportional performance gains. The self-citation to the dissertation study is included because the present work reuses its repeated-measures token-scaling structure and extends its parameter-efficiency logic by implementing the energy-efficiency direction recommended in Chapter 5.

\section{Methods}
This study employed a repeated-measures experimental design to evaluate the effect of training token count on energy-aware parameter efficiency under fixed hardware, software, and optimization conditions. A within-subject design was selected to control for variance attributable to model architecture, hardware configuration, and training hyperparameters, thereby isolating training token count as the sole manipulated variable.

The experimental design was adapted from the dissertation study, which used the same broad token-count conditions and the same conceptual parameter-efficiency framework (Dwyer, 2025). The present study differs from the earlier work by adding direct GPU power monitoring and by incorporating RMS power consumption into the dependent efficiency metric. This extension allows each trial to be evaluated in terms of both model behavior and measured energy cost.

Three training token conditions were evaluated: 500,000, 1,000,000, and 2,000,000 true tokens. For each condition, 50 training trials were conducted, yielding a total of 150 trials. Each trial initialized a fresh instance of the model and dataset to preserve independence across runs. No curriculum learning, adaptive scheduling, or early stopping mechanisms were employed. All experiments were executed on the same hardware instance and under identical software configurations.

All experiments utilized the TinyLlama 1.1B Chat v1.0 causal language model architecture, which has approximately 1.1 billion parameters. Training was performed using the AdamW optimizer with a fixed learning rate of $1 \times 10^{-6}$. Mixed precision FP16 training was employed using automatic mixed precision with gradient scaling (Dwyer, 2026). Gradient norms were clipped at 1.0. Each trial executed three training epochs. Individual batches were skipped only if the loss became nonfinite or exceeded the predefined stability threshold used in the training script (Dwyer, 2026). This safeguard prevented unstable gradient updates from contaminating the run while preserving the fixed three-epoch structure. As a result, the number of optimizer updates could vary slightly across trials, but the number of epochs remained fixed.

Training data were drawn from a JSONL-formatted TinyStories corpus (Dwyer, 2025; Dwyer, 2026). Tokenization employed fixed-length padding and truncation, with a maximum sequence length of 64 tokens. Padding was applied using the end-of-sequence token. Dataset construction proceeded incrementally until the target number of true non-padding tokens, determined via the attention mask, was reached or slightly exceeded. The batch size was fixed at one, and the sample order was randomized.

All experiments were conducted on an AWS SageMaker AI \emph{ml.g5.xlarge} instance equipped with a single NVIDIA A10G GPU (Dwyer, 2025; Dwyer, 2026). GPU power consumption was monitored using the NVIDIA Management Library. Power readings were sampled every 60 seconds via a background monitoring thread, with an immediate initial sample collected at startup (Dwyer, 2026). Power values were recorded in watts and aggregated using the root mean square of all samples collected during each trial.

Model performance was evaluated using inverse perplexity, computed from the post-training evaluation loss. Evaluation was initially attempted using a fixed default prompt, followed by fallback prompts when a prompt failed to produce a numerically stable evaluation loss (Dwyer, 2026). In rare cases, when prompt-based evaluation did not return a stable value, perplexity was recovered from a valid training batch to avoid discarding an otherwise completed trial. These cases were handled using the same numerical stability rules as the primary evaluation procedure: perplexity was computed as the exponential of loss only when the loss value was finite and numerically stable, and inverse perplexity was defined as its reciprocal. This procedure ensured that each retained trial had a valid performance signal while making the fallback logic explicit for replication.

The primary dependent variable was parameter efficiency, computed using an energy-aware formulation integrating inverse perplexity, benchmarked compute throughput, measured power consumption, model size, and total training tokens processed. Compute throughput was represented by a fixed benchmark TFLOPS constant selected as the higher of empirically measured FP16 and BF16 values (Dwyer, 2025; Dwyer, 2026). Power efficiency was expressed as TFLOPS per watt using RMS power measurements. The total tokens processed were computed as the product of the target token count and the epoch count, normalized to millions of tokens. An additional derived metric, parameter efficiency loss, was computed relative to a fixed baseline.

Statistical analyses were conducted using Python and the \emph{Pingouin} library (Dwyer, 2025; Dwyer, 2026). A one-factor repeated measures ANOVA was performed to assess the effect of training token count on parameter efficiency. Normality was assessed using the Shapiro-Wilk test. Sphericity was evaluated using Mauchly's test, with conservative corrections planned if violations were detected. Post hoc pairwise comparisons were performed using the Bonferroni correction for multiple comparisons. Effect size was reported using generalized eta squared ($\eta_g^2$). Statistical significance was evaluated at an $\alpha$ level of 0.05.

\section{Results}
All 150 training trials were completed successfully and retained for analysis, with each trial producing at least one NVIDIA Management Library power sample (Dwyer, 2026). Descriptive statistics revealed a monotonic decline in parameter efficiency as training token count increased, accompanied by increased power consumption, longer execution duration as reflected by power sampling counts, and increased perplexity per parameter.

A one-factor repeated measures ANOVA revealed a statistically significant effect of training token count on parameter efficiency, F(2, 98) = 24,268.23, $p < .001$, $\eta_g^2 = .997$. Bonferroni-corrected pairwise comparisons indicated significant differences between all token conditions. Assumption testing indicated that normality and sphericity assumptions were satisfied.

\begin{table}[htbp]
    \centering
    \caption{Bonferroni Corrections for Parameter Efficiency}
    \begin{tabular}{ccccc}
        \hline
        Token Count & $t$ value & Raw $p$ Value & Corrected $p$ & Reject $H_0$ \\
        \hline
        500K vs 1M & 117.9388 & 0.00 & 0.00 & True \\
        500K vs 2M & 220.8192 & 0.00 & 0.00 & True \\
        1M vs 2M   & 97.6873  & 0.00 & 0.00 & True \\
        \hline
    \end{tabular}
\end{table}

\begin{table}[htbp]
    \centering
    \caption{Shapiro-Wilk Normality Test for Parameter Efficiency}
    \begin{tabular}{cccc}
         \hline
         Token Count & W value & $p$ value & Result \\
         \hline
         500K & 0.9755 & 0.3796 & Normal \\
         1M & 0.9735 & 0.3200 & Normal \\
         2M & 0.9880 & 0.8871 & Normal \\
         \hline
    \end{tabular}
\end{table}

To determine whether the training token count independently influenced energy consumption, a one-factor repeated measures analysis of variance was conducted on the root mean square GPU power usage measured during training. Power consumption was aggregated per trial using RMS values derived from NVIDIA Management Library samples collected at regular intervals throughout each training run (Dwyer, 2026).

The repeated measures ANOVA revealed a statistically significant main effect of training token count on power consumption, F(2, 98) = 3143.21, $p < .001$. The generalized eta squared effect indicated a substantial effect, $\eta_g^2 = .969$, demonstrating that the majority of within-subject variance in power consumption was attributable to differences in training token count.

Bonferroni-corrected pairwise comparisons revealed statistically significant differences between all token count conditions. Mean RMS power consumption in the 500,000 token condition differed significantly from the 1,000,000 token condition, $t = -40.69$, $p < .001$, and from the 2,000,000 token condition, $t = -84.16$, $p < .001$. The 1,000,000-token condition also differed significantly from the 2,000,000-token condition, $t = -36.65$, $p < .001$. Across conditions, mean RMS power consumption increased monotonically with training token count, reflecting sustained increases in energy usage under otherwise fixed training parameters.

\begin{table}[htbp]
    \centering
    \caption{Bonferroni Corrections for Power Consumption}
    \begin{tabular}{ccccc}
        \hline
        Token Count & $t$ value & Raw $p$ Value & Corrected $p$ & Reject $H_0$ \\
        \hline
        500K vs 1M & -40.6892 & 0.00 & 0.00 & True \\
        500K vs 2M & -84.1646 & 0.00 & 0.00 & True \\
        1M vs 2M   & -36.6477  & 0.00 & 0.00 & True \\
        \hline
    \end{tabular}
\end{table}

Assumption testing indicated that power consumption values were normally distributed for the 500,000 token condition but deviated from normality for the 1,000,000 and 2,000,000 token conditions. Mauchly's test indicated a violation of the sphericity assumption, W = 0.422, $\chi^2(2) = 41.43$, $p < .001$. Accordingly, Greenhouse--Geisser corrected results were examined and remained statistically significant ($p < .001$), confirming the robustness of the observed effect.

\begin{table}[htbp]
    \centering
    \caption{Shapiro-Wilk Normality Test for Power Consumption}
    \begin{tabular}{cccc}
         \hline
         Token Count & W value & $p$ value & Result \\
         \hline
         500K & 0.9799 & 0.5475 & Normal \\
         1M & 0.8175 & 0.00 & Non-normal \\
         2M & 0.9487 & 0.0300 & Non-normal \\
         \hline
    \end{tabular}
\end{table}

These results demonstrate that increases in training token count produce significant and systematic increases in energy consumption under fixed hardware, optimizer configuration, and epoch count. Importantly, this effect was observed independently of model performance outcomes and provides direct empirical evidence that token scaling imposes substantial energy costs even when architectural and algorithmic variables are held constant.

\section{Discussion}
The results demonstrate that increasing training token count under fixed conditions produces predictable and substantial reductions in energy-aware parameter efficiency. Despite statistically significant separation between conditions, inverse perplexity varied only modestly, indicating that efficiency losses were driven primarily by increased computational and energy cost rather than degraded model behavior. These findings extend prior scaling law work by showing how efficiency-based evaluation alters conclusions when energy is treated as a first-class constraint.

The additional analysis of GPU power consumption clarifies the mechanism underlying the observed decline in energy-aware parameter efficiency. While prior results demonstrated that increasing the training token count produced large and statistically significant reductions in efficiency, the power-focused repeated measures analysis shows that these efficiency losses are driven in part by substantial increases in sustained energy usage. Power consumption increased monotonically with the token count and exhibited a significant within-subject effect, even under fixed hardware, optimizer configuration, and epoch count. Notably, these increases in energy demand occurred alongside relatively modest changes in inverse perplexity, indicating that the efficiency collapse observed at higher token counts is not attributable to degraded model behavior but rather to disproportionate growth in computational cost.

This finding reinforces the central claim of the study: when energy consumption is treated as a first-class variable, training token scaling reveals predictable and substantial inefficiencies that are obscured by performance-centric evaluation alone. Taken together, the efficiency and power analyses demonstrate that token count operates as a direct driver of energy cost under controlled conditions, providing empirical support for efficiency-aware interpretations of scaling behavior and highlighting the need to integrate power measurements into evaluations of large language model training strategies.

This study also clarifies how practitioners should interpret statistically significant efficiency results. Statistical separation between token-count conditions demonstrates that token count affects the measured outcome, but it does not automatically imply that larger token counts produce practically desirable results. For constrained training settings, the appropriate interpretation requires both statistical testing and performance-cost interpretation. Researchers and practitioners should therefore evaluate token-scaling decisions using a dual lens: whether a statistically detectable effect exists, and whether the resulting efficiency tradeoff is acceptable given available hardware, energy, budget, and sustainability constraints.

\section{Limitations and Future Work}
This study is limited to a single model architecture and GPU class, with all experiments conducted on a 1.1-billion-parameter TinyLlama model trained on an NVIDIA A10G AWS instance. This constraint was deliberately imposed to isolate the effects of training token count and energy consumption without confounding architectural or hardware variability. As a result, while the specific quantitative outcomes reported here may not directly transfer to larger models or alternative accelerator types, the proposed energy-aware parameter efficiency formulation is not inherently tied to a particular architecture or device. Future work should extend this framework to additional model families, parameter scales, and hardware platforms, including multi-GPU and distributed training settings, to evaluate the generality of the observed efficiency dynamics under more heterogeneous conditions.

Future research should also combine the two follow-up directions identified in the dissertation: more granular token intervals and energy-aware efficiency metrics. The present study implements the energy-measurement direction, but it retains the original broad token-count intervals of 500K, 1M, and 2M. Additional experiments using intermediate token counts, such as 1.1M, 1.2M, and subsequent intervals up to 2M, could help identify whether the observed decline reflects a smooth non-linear pattern, an inflection point, or a threshold effect. Combining granular token variation with direct power measurement would strengthen the practical value of the metric for researchers selecting token budgets under constrained conditions.

\section{Conclusion}
This study provides a confirmatory replication and extension of token scaling behavior under energy-aware constraints. Increasing the training token count, while holding model architecture, hardware, optimizer configuration, and epoch count constant, produces systematic and predictable reductions in parameter efficiency. These results reinforce the concept of diminishing returns in neural scaling, while also highlighting the need to integrate direct energy measurement into training evaluation.

The practical implication is that researchers should avoid treating token count as an automatically beneficial scaling lever. Instead, token budgets should be empirically tuned and evaluated with metrics that account for both performance and resource cost. Reporting power consumption, execution duration, and energy-aware efficiency alongside conventional loss or perplexity metrics would make training studies more reproducible, more transparent, and more useful for researchers operating outside high-resource industrial laboratories.

By implementing the dissertation's recommendation to incorporate parameter efficiency per watt, this study moves the framework from compute-aware evaluation toward energy-aware evaluation. Efficiency-aware analysis provides a more comprehensive and practically relevant framework for evaluating large language model training strategies, particularly in constrained and sustainability-focused contexts. These findings suggest that sustainable LLM development will require not only larger models or larger datasets, but more careful measurement of when additional training scale stops producing proportionate value.

\nocite{*}
\bibliographystyle{apacite}
\bibliography{references}

\end{document}